\documentclass{IEEEtran}
\usepackage{graphicx}
\usepackage{amsmath}
\usepackage{cite}
\usepackage{booktabs}
\usepackage{multirow}
\usepackage{xcolor,colortbl}
\usepackage{svg}
\usepackage{amsfonts}
\usepackage{amssymb}
\usepackage{arydshln}
\usepackage{verbatim}

\title{ExIQA: Explainable Image Quality Assessment Using  Distortion Attributes}
\author{
\vspace{0.5cm}

{\Large Sepehr Kazemi Ranjbar $\quad$ Emad Fatemizadeh
\\
\vspace{0.25cm}
Department of Electrical Engineering, Sharif University of Technology\\
Tehran, Iran}
\\
\vspace{0.25cm}
\{sepehr.kazemi99, fatemizadeh\}@sharif.edu}

\newtheorem{definition}{Definition}

\newcommand{\gray}[1]{{\color{gray} #1}}

\usepackage[pagebackref,breaklinks,colorlinks]{hyperref}

\usepackage[capitalize]{cleveref}
\crefname{section}{Sec.}{Secs.}
\Crefname{section}{Section}{Sections}
\Crefname{table}{Table}{Tables}
\crefname{table}{Tab.}{Tabs.}

\begin{document}
\maketitle

\begin{abstract}
Blind Image Quality Assessment (BIQA) aims to develop methods that estimate the quality scores of images in the absence of a reference image. In this paper, we approach BIQA from a distortion identification perspective, where our primary goal is to predict distortion types and strengths using Vision-Language Models (VLMs), such as CLIP, due to their extensive knowledge and generalizability. Based on these predicted distortions, we then estimate the quality score of the image. To achieve this, we propose an explainable approach for distortion identification based on attribute learning. Instead of prompting VLMs with the names of distortions, we prompt them with the attributes or effects of distortions and aggregate this information to infer the distortion strength. Additionally, we consider multiple distortions per image, making our method more scalable. To support this, we generate a dataset consisting of 100,000 images for efficient training. Finally, attribute probabilities are retrieved and fed into a regressor to predict the image quality score. The results show that our approach, besides its explainability and transparency, achieves state-of-the-art (SOTA) performance across multiple datasets in both PLCC and SRCC metrics. Moreover, the zero-shot results demonstrate the generalizability of the proposed approach.\footnote{Codes and datasets will be available online upon publication}   
\end{abstract}

\begin{figure*}[t]
    \centering
    \includegraphics[width=0.8\linewidth]{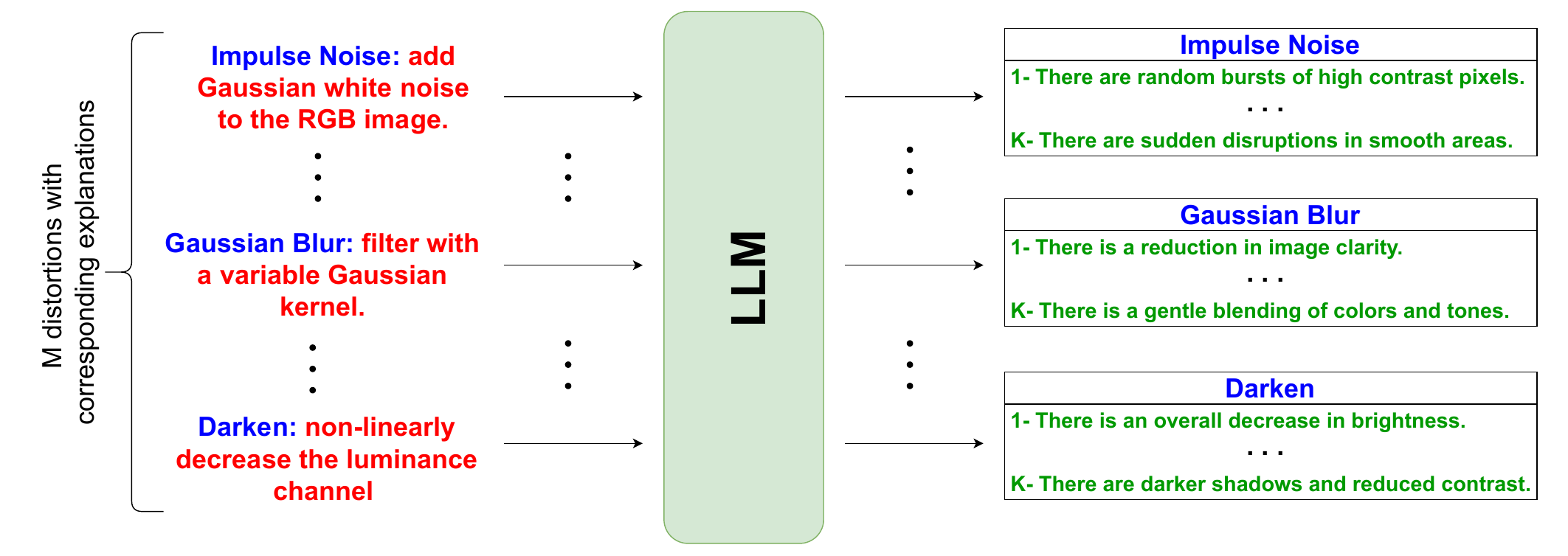}
    \caption{Use of LLMs, such as GPT-3, to generate visual effects/attributes of distortions. Each distortion with an explanation is fed to the LLM. In output, we have $K$ visual attributes for that distortion.}
    \label{fig:1}
\end{figure*}

\section{Introduction}\label{sec:intro}
Blind Image Quality Assessment (BIQA) is the task of predicting the quality scores of a degraded image in the absence of its pristine counterpart \cite{wang2006modern}. Deep Learning models address this task by training a large number of parameters with limited annotated datasets \cite{zhai2020perceptual, bosse2017deep, athar2019comprehensive}, making them prone to overfitting. Additionally, their lack of explainability limits their application in critical domains such as medical imaging \cite{chow2016review}.

To enhance transparency and improve the performance of Blind Image Quality Assessment (BIQA), we consider a related task: distortion identification \cite{liang2020deep, yan2021precise}, where the goal is to determine the type and strength of distortions in a photo. The identified distortions are then used as input features for a regressor to estimate the image’s quality score. Since we use only distortions as input features for the regressor, the quality score is predicted transparently without relying on irrelevant features that may compromise explainability and generalizability. However, the key challenge lies in how to predict distortions in an explainable manner. This paper employs Vision-Language Models (VLMs) \cite{zhang2024vision, zhang2021vinvl}, such as CLIP \cite{radford2021learning}, as explainable distortion identifiers.

Some prior works have used large vision (-language) models as distortion identifiers. A relevant study by Agnolucci et al. \cite{arniqa} fine-tuned a pre-trained ResNet-50 \cite{chen2020simple} as an image encoder, then retrieved features from the image encoder to train a regressor. They employed a similarity loss for crops from different images but with the same distortions and a dissimilarity loss for crops of the same image. This method guides the model to shift its focus from content to distortions, which is essential. However, incorporating text modality could enhance the decision-making process and make the model more explainable. Additionally, they fed the regressor with all the features extracted from the image encoder, which impacts the model's transparency, as there is no guarantee that all features are relevant, increasing the risk of shortcut learning \cite{ma2023rectify}.

Another work closely related to ours is by Zhang et al. \cite{zhang2023blind}, who used a multi-task learning framework and fed the text encoder the following compound text prompt: {\small\fontfamily{qcr}\selectfont "a photo of a <\textbf{category}> with <\textbf{distortion type}> artifacts, which is of <\textbf{score}> quality"}. The main limitation of their approach is the use of exact distortion names in the prompt, which restricts scalability to variant distortion types, including those with ambiguous names. Consequently, they only consider ten distortion types. Another drawback is that the network is not trained to identify distortion strengths in the photo, which is crucial for estimating the image's quality score. Furthermore, they consider only a single distortion, limiting the model's representational capability.

In this paper, we address the distortion identification problem by focusing on the effects of distortions rather than relying on specific names. In other words, we feed the text encoder with prompts containing attributes (or effects) of distortions extracted from experts or Large Language Models (LLMs) \cite{zhao2023survey} such as GPT-3 \cite{brown2020language} for a more automated process. We then aggregate the likelihood of these attributes to infer the distortion strength. This approach achieves three critical objectives simultaneously: first, since we do not depend solely on ambiguous names and instead consider the natural effects of distortions, our method is explainable. Second, our model is scalable to any type and number of distortions, provided the effects of those distortions are known. Third, we enable the model to predict not only the distortion type but also the distortion strength as a continuous variable, enhancing the model’s representation. Additionally, our approach accounts for multiple distortions per image. To support this, we generated a large dataset of 100k images based on the KADID-10k dataset \cite{kadid10k}, where each image contains multiple distortions. Finally, to predict the quality score of the images, we trained a regressor that only uses attribute probabilities as input features. This prevents the inclusion of irrelevant features that could lead to intransparency and shortcut learning in the regressor network. The results demonstrate that we achieved state-of-the-art (SOTA) performance across different datasets and domains. We also observed promising zero-shot performance, highlighting the generalizability of our approach. In summary, our contributions are as follows: 
\begin{itemize} 
\item Designing a novel approach for distortion identification based on distortion's effects/attributes
\item Accounting multiple distortions for each image and generating a large dataset of 100,000 multi-distorted images
\item Using only attribute probabilities as input features for regressor network and achieving SOTA performance across different datasets besides computational efficiency.
\end{itemize}

The structure of the paper is as follows: In Section 2, we review the literature on works related to our task. Section 3 discusses some preliminary concepts, such as CLIP and prompt-tuning. In Section 4, we define our problem mathematically and propose our approach. In Section 5, we conduct multiple experiments to evaluate our method. Finally, Section 6 concludes the paper and discusses potential future works.

\section{Related Work}\label{sec:related_work}
\subsection{Blind Image Quality Assessment}
 Classical BIQA methods, such as NIQE \cite{niqe} and BRISQUE \cite{brisque}, heavily rely on handcrafted features \cite{fu2018quality} and natural scene statistics \cite{reinagel1999natural}. Codebook-based methods \cite{ye2012no}, like CORINA \cite{cornia} and HOSA \cite{hosa}, use visual codebooks to identify patterns among image patches and then determine the image's quality score. With the rapid progress of deep learning across various fields, methods like DB-CNN \cite{dbcnn}, DeepFL-IQA \cite{lin2020deepfl}, and DeepBIQ \cite{bianco2018use} have emerged, outperforming classical methods by leveraging the high feature extraction capability of Convolutional Neural Networks. However, the main limitation of these deep learning methods is their lack of explainability \cite{angelov2020towards} and reduced generalizability \cite{zhang2019deep} when applied to different domains and datasets.
Self-supervised learning methods, such as ARNIQA \cite{arniqa}, Re-IQA \cite{reiqa}, CONTRIQUE\cite{contrique}, and VISOR\cite{zhou2023collaborative}, partially address this issue by forcing the backbone network to extract distortion-based features. Recently, with the advent of Vision-Language Models (VLMs), some methods, such as LIQE \cite{zhang2023blind}, which is the most relevant to our work, have used language prompts to detect distortions alongside predicting the image quality score. However, due to the limited number of distortion names recognizable by VLMs, these methods still face challenges in scalable distortion identification, which impacts the model’s representation and quality score prediction.

\subsection{Visual Attributes Reasoning}
Our work is inspired by studies that use visual attributes (rationales) for image classification \cite{mao2023doubly, rasekh2024ecor}. Similarly, in the commonsense reasoning task \cite{zellers2019recognition, davis2015commonsense}, interpretations are often made to predict future states of the world or the actions of people or objects based on current observations. In this work, we apply this idea to predict distortions by analyzing their attributes. We focus on understanding the attributes or effects caused by distortions, with the objective of finding an accurate mapping between different attributes and various distortions.

\begin{figure*}[t]
    \centering
    \includegraphics[width=\linewidth]{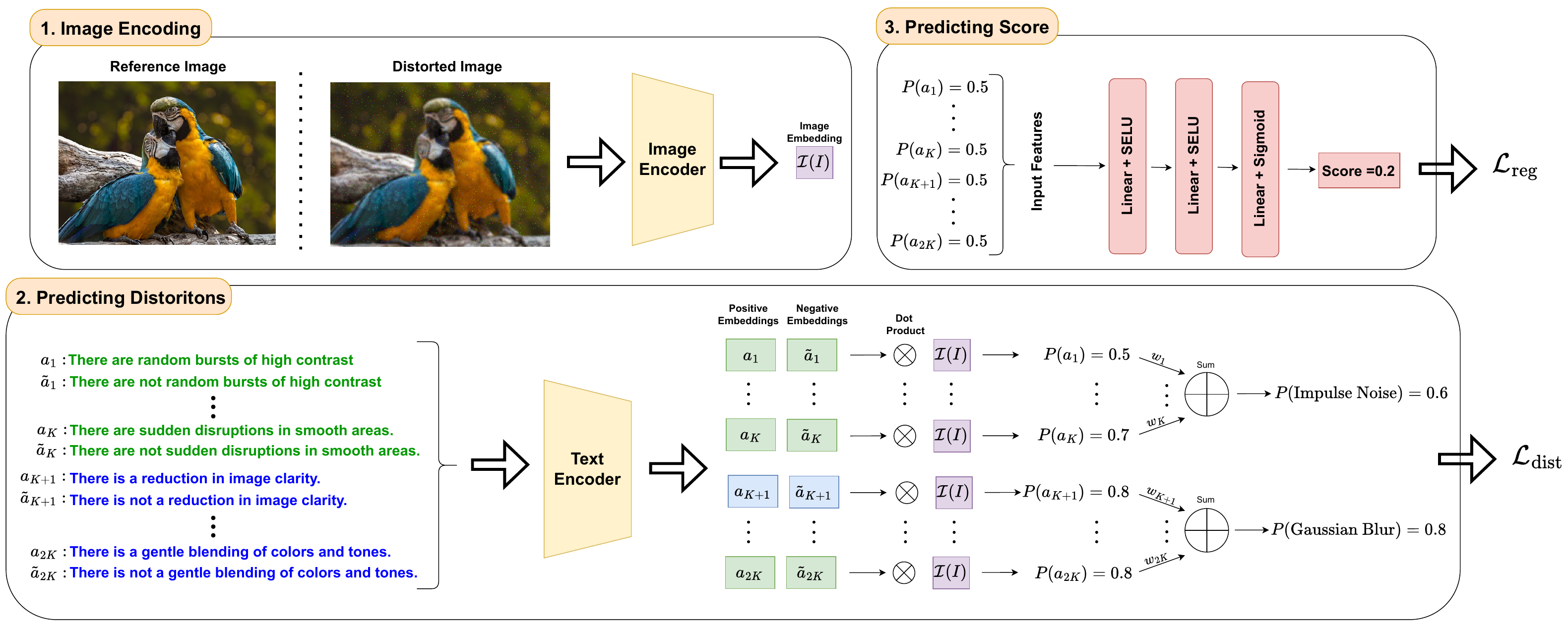}
    \caption{The architecture of our model. We consider two distortions for visual simplicity: Impulse Noise and Gaussian Blur. The image also has both of them with strengths of 0.6 and 0.8. The reference image is shown just for comparison with the distorted image. The pipeline is as follows: 1) The distorted image is encoded. 2) The positive and negative prompts for all attributes are generated (two distortions, each has $K$ attributes). Then, the probability of each attribute is computed. Finally, for each distortion, its $K$ attribute probabilities are averaged to obtain the distortion probability. 3) Attributes probabilities are retrieved and fed to the regressor network for prediction of the quality score.}
    \label{fig:2}
\end{figure*}

\section{Preliminaries}
\subsection{CLIP Overview}\label{subsec:clip}
CLIP (Contrastive Language-Image Pre-Training) \cite{radford2021learning} is a Vision Language Model consists of a text encoder, $\mathbf{E}_{T}:\text{text}\rightarrow \mathbb{R}^{d}$ and an image encoder $\mathbf{E}_{I}:\text{image}\rightarrow \mathbb{R}^{d}$ where $d$ is the multi-modal space dimension of CLIP. It was trained on 400 million (image-caption) pairs, i.e., $\left\{I_i, t_i\right\}_{i=1}^{N}$, via the following contrastive loss:

\begin{align}
    \mathcal{L}_{\text{CLIP}} &= -\frac{1}{N}\sum_{i=1}^{N} \log\left( \frac{e^{\mathbf{E}_I(I_i)^{T}\mathbf{E}_{T}(t_i)}}{\sum_{j=1}^{N}e^{\mathbf{E}_I(I_i)^{T}\mathbf{E}_{T}(t_j)}}\right)\nonumber\\
    &- 
    \frac{1}{N}\sum_{i=1}^{N}\log \left(\frac{e^{\mathbf{E}_I(I_i)^{T}\mathbf{E}_{T}(t_i)}}{\sum_{j=1}^{N}e^{\mathbf{E}_I(I_j)^{T}\mathbf{E}_{T}(t_i)}}\right)
\end{align}

For inference, when we have an image $I$ and set of candidate captions $\left\{t_i\right\}_{i=1}^{M}$, the most proper caption, $\hat{t}$ is given by:
\begin{align}
    \hat{t} = \text{argmax}_{t_i}\quad \mathbf{E}_{I}(I)^T \mathbf{E}_{T}(t_i)
\end{align}

\subsection{Prompt Tuning}\label{subsec:pt}
To save the generalizability of CLIP and reduce training time, we utilize prompt-tuning methods for fine-tuning CLIP. In particular, we consider two prompt-tuning methods, shallow-prompt tuning \cite{mao2023doubly} and deep-prompt tuning\cite{jia2022visual}.
\subsubsection{Shallow-Prompt Tuning}
In this case, $K$ learnable prompts are concatenated to image tokens at the vision transformer (image encoder) input:
\begin{align}
    \text{input} = \left[x,p_1,p_2,\dots,p_K\right]
\end{align}
where $x\in \mathbb{R}^{H\times d}$ is the image tokens ($H$ is number of tokens), and $\left\{p_i\in \mathbb{R}^{d}\right\}_{i=1}^{K}$ are learnable prompts.

\subsubsection{Deep-Prompt Tuning}
The shallow-prompt tuning can be extended to all layers of the vision transformer:
\begin{align}
    x^{(l)} &= \left[y^{(l-1)},p^{(l)}_{1},p^{(l)}_{2},\dots,,p^{(l)}_{K}\right]\\
    y^{(l)} &= \text{Transformer}^{(l)}(x^{(l)})\left[1:H\right]
\end{align}
Where $y^{(l)}$ is the output of $l$th layer after removing the corresponding prompts appended at the layer's input. $\text{Transformer}^{(l)}$ is the transformer block at layer $l$ containing multi-head-attention \cite{vaswani2017attention}, LayerNorm\cite{ba2016layer} and MLP\cite{lecun2015deep}.

\section{Proposed Model} \label{sec:method}
\subsection{Problem Formulation}\label{subsec:PF}
Consider a dataset $\zeta = (\mathcal{I},\, \mathcal{D},\, \mathcal{P},\, \mathcal{S})$, where $\mathcal{I}$ is the set of images, $\mathcal{D}$ is the set of distortions, $\mathcal{P}:\mathcal{I}\times\mathcal{D}\rightarrow \mathbb{R}$ is the distortion strength function, and $\mathcal{S}:\mathcal{I}\rightarrow \mathbb{R}$ is the quality score function. For each image $I\in \mathcal{I}$ and every $d\in \mathcal{D}$, there exists $\mathcal{P}(I, d)\in \left[0,1\right]$, which determines the probability\footnote{In this paper, we use the terms "strength" and "probability" interchangeably.} or strength of distortion $d$ in image $I$. If image $I$ has not distortion $d$, then $\mathcal{P}(I,d)=0$. Additionally, $\mathcal{S}(I)\in \left[0,1\right]$ represents the normalized quality score of the image, where a higher score indicates better quality.

We define our problem as the following two sub-problems.

\begin{definition}[Sub-Problem 1]
    Consider dataset $\zeta$, given image $I$, we aim to find $\mathcal{P}(I,d)$ for all $d\in \mathcal{D}$.
\end{definition}

\begin{definition}[Sub-Problem 2]
    Given image $I$, we aim to predict its quality score, i.e., $\mathcal{S(I)}$, using predicted distortions.
\end{definition}

\begin{table*}[t]
    \small
    \centering
        \caption{Datasets statistics}
    \begin{tabular}{llllll}
         \toprule
         Dataset & Type & \begin{tabular}{l} No. of \\Distortions\end{tabular} & \begin{tabular}{l} No. of \\ Reference Images\end{tabular} & \begin{tabular}{l} No. of \\ Distorted Images\end{tabular} & Score Range\\
        \midrule
        KADID-10k \cite{kadid10k} &  single-distortion & 25 & 81 & 10,125 & $[1,5]$\\
        TID2013 \cite{tid2013}&  single-distortion & 24 & 25 & 3,000 & $[0,9]$\\
        CSIQ \cite{csiq} &  single-distortion & 6 & 30 & 866 & $[0,1]$\\
        LIVE \cite{live}&  single-distortion & 5 & 29 & 779 & $[1,100]$\\
        LVQ \cite{lvq1}&  single-distortion & 5 & 2561 (frames) & 51220 (frames) & $[0,3]$\\
        \textbf{ExIQA (ours)} & multi-distortion & 25 & 10,000 & 100,000 & -\\
        \bottomrule
    \end{tabular}
    \label{tabel:ds}
\end{table*}

\begin{table}[t]
    \small
    \centering
    \caption{Results of distortions estimation. The accuracy and RMSE are averaged among all distortions in each dataset.}
    \begin{tabular}{llllll}
         \toprule
         Dataset & \begin{tabular}{l} Prompt-Tuning Type \\ \& Length \end{tabular} & Accuracy & RMSE\\
        \midrule
        KADID-10k & Shallow \& 100 & 97.65 & 0.0347\\
        TID2013 & Shallow \& 100  & 96.98 & 0.0581\\
        CSIQ & Shallow \& 15 & 92.44 & 0.0572\\
        LVQ & Shallow \& 30  & 98.37 & 0.0315\\
        LIVE & Shallow \& 15 & 93.42 & 0.0697\\ 
        ExIQA & Whole CLIP & 68.82 & 0.156\\
        \bottomrule
    \end{tabular}
    \label{tabel:m1}
\end{table}

\subsection{Identifying Distortions with their Attributes}
For predicting distortions via VLMs such as CLIP, previous methods \cite{zhang2023blind} used distortion names in their textual prompts. However, this approach is not reliable because the names of distortions can be too ambiguous or complex for the CLIP text encoder to understand effectively. For example, consider the distortion \emph{"Color saturation 1"} in the KADID-10K dataset \cite{kadid10k}, which is created by multiplying the saturation channel in the HSV color space by a factor. As seen, the name of this distortion is ambiguous, reducing the trustworthiness of the model's prediction.

In this paper, we address this issue by focusing on the effects of distortions instead of their names. For instance, Gaussian Blur causes image edges and overall texture to appear smoother. Based on the strength of these effects, one can estimate the distortion probability. To achieve this, instead of using the distortion's name as a text prompt, we use the visual effects or attributes of the distortion as text prompts, allowing the VLM to identify these visual attributes in the image. This approach makes our distortion identification more explainable and accurate.

Although a single attribute may be shared by multiple distortions, we consider multiple attributes for each distortion to distinguish them effectively.

\subsubsection{LLM for find attributes}
One way to identify distortion attributes is through human experts, which is time-consuming. However, we propose a more automated and scalable approach using LLMs.

LLMs, such as GPT-3 \cite{brown2020language}, are trained on billions of corpora and generally possess far more knowledge than VLMs like CLIP \cite{radford2021learning}. Due to their reasoning capabilities \cite{espejel2023gpt}, they can generate visual effects associated with distortions.

For each distortion $d \in \mathcal{D}$, we provide a textual explanation detailing how this distortion is applied to an image. We query GPT-3 with the distortion name and its corresponding explanation, using a specific prompt to generate $K$ (a hyperparameter) visual attributes for the given distortion. Figure \ref{fig:1} illustrates this procedure. Below is an example of the prompt provided to GPT-3, where we aim to extract $5$ attributes for the Gaussian Blur \cite{kadid10k} distortion.
\\
\\
Input to GPT-3:
\\
{\small\fontfamily{qcr}\selectfont
Give me five visual attributes that describe the visual effects of the given distortion.
\\
\\
\#INPUT
\\
Gaussian blur: filter with a variable Gaussian kernel.
\\
\\
}
Output of GPT-3:
{\small\fontfamily{qcr}\selectfont
\\
1- There is a softening of details in the photo.
\\
2- There is a gentle blending of colors and tones.
\\
3- There is a reduction in image clarity.
\\
4- There is a slight loss of fine texture, edges, and detail.
\\
5- There is a dreamy or softened appearance to the overall image.}

\begin{table*}[t]
    \small
    \centering
    \caption{Results on synthetic datasets. The bold results in each column are the best, and the underlined results are the second best. For LVQ dataset, due to the lack of evaluation on some baselines, we train and evaluate with the code in their repository, so the gray results are evaluated by us. As can be seen, we got SOTA performance in SRCC metric across all datasets.}
    \begin{tabular}{*{13}{l}}
    \toprule 
        & \multicolumn{2}{c}{KADID-10k} & \multicolumn{2}{c}{TID2013} & \multicolumn{2}{c}{CSIQ} & \multicolumn{2}{c}{LVQ} & \multicolumn{2}{c}{LIVE} & \multicolumn{2}{c}{Average} \\
        
         Method & PLCC & SRCC & PLCC & SRCC & PLCC & SRCC & PLCC & SRCC & PLCC & SRCC &  PLCC & SRCC\\
        \midrule
         NIQE \cite{niqe} & 0.428 & 0.374 & 0.393 & 0.315 & 0.712 & 0.627 & 0.458 & 0.499 &  0.901 & 0.907 & 0.578 & 0.544\\
         BRISQUE \cite{brisque} & 0.567 & 0.528 & 0.694 & 0.604 & 0.829 & 0.746 & 0.472 & 0.436 & 0.935 & 0.939 & 0.699 & 0.651\\
         CORNIA \cite{cornia}& 0.558 & 0.516 & 0.768 & 0.678 & 0.776 & 0.678 & \gray{0.567} & \gray{0.519} & 0.950 & 0.947 & 0.724 & 0.668\\
         HOSA \cite{hosa}& 0.653 & 0.618 & 0.815 & 0.735 & 0.823 & 0.741 & \gray{0.602} & \gray{0.577} & 0.950 & 0.946 & 0.769 & 0.723\\
         DB-CNN \cite{dbcnn}& 0.856 & 0.851 & 0.865 & 0.816 & 0.959 & 0.946 & \gray{0.902} & \gray{0.876} & \textbf{0.971} & 0.968 & 0.911 & 0.891\\
         HyperIQA \cite{hyperiqa}& 0.845 & 0.852 & 0.858 & 0.840 & 0.942 & 0.923 & \gray{0.937} & \gray{0.911} & 0.966 & 0.962 & 0.910 & 0.898\\
         TReS \cite{tres}& 0.858 & 0.859 & 0.883 & 0.863 & 0.942 & 0.922 & \gray{0.928} & \gray{0.918} & 0.968 & 0.969 & 0.916 & 0.907\\
         CONTRIQUE \cite{contrique}& \underline{0.937} & \underline{0.934} & 0.857 & 0.843 & 0.955 & 0.942 & \gray{0.959} & \gray{0.945} & 0.961 & 0.960 & 0.934 & 0.925\\
         Re-IQA \cite{reiqa}& 0.885 & 0.872 & 0.861 & 0.804 & 0.960 & 0.947 & \gray{0.961} & \underline{\gray{0.958}} & \textbf{0.971} & \underline{0.970} & 0.928 & 0.910\\
         LIQE \cite{zhang2023blind} & 0.931 & 0.930 & \gray{0.875} & \gray{0.856} & 0.939 & 0.936 & \gray{0.969} & \gray{0.944} & 0.951 & 0.970 & 0.933 & 0.927\\
         
         ARNIQA \cite{arniqa}& 0.912 & 0.908 & \textbf{0.901} & \underline{0.880} & \textbf{0.973} & \underline{0.962} & \underline{\gray{0.970}} & \gray{0.937} & \underline{0.970} & 0.966 & \underline{0.945} & \underline{0.931}\\
         \midrule
         \textbf{ExIQA (ours)} & \textbf{0.943} & \textbf{0.939} & \underline{0.899} & \textbf{0.912} & \underline{0.961} & \textbf{0.965} & \textbf{0.987} & \textbf{0.966} & 0.967 & \textbf{0.973} & \textbf{0.951} & \textbf{0.951}\\
         \bottomrule
    \end{tabular}
    \label{tabel:m2}
\end{table*}

\begin{table}[t]
    \small
    \centering
        \caption{Zero-Shot result for distortion identification task.}

    \begin{tabular}{llll}
         \toprule
         \begin{tabular}{l} Training \\ Dataset \end{tabular} & 
         \begin{tabular}{l} Testing \\ Dataset \end{tabular} & 
         Accuracy & RMSE \\
         \midrule
         KADID-10k & TID2013 & 81.15 & 0.1612\\
         KADID-10k & CSIQ & 71.41 & 0.2710\\
         KADID-10k & LVQ & 84 & 0.2710\\
         KADID-10k & LIVE & 70.81 & 0.1914\\
        \midrule
         TID2013 & KADID-10k & 88.51 & 14.49\\
         TID2013 & CSIQ & 66.41 & 0.2528\\
         TID2013 & LVQ & 83.94 & 0.2722\\
         TID2013 & LIVE & 72.33 & 0.1893\\
        \midrule
         ExIQA & KADID-10k & 94.00 & 0.1285\\
         ExIQA & TID2013 & 90.35 & 0.1421\\
         ExIQA & CSIQ & 76.15 & 0.2383\\
         ExIQA & LVQ & 88.97 & 0.2382\\
         ExIQA & LIVE & 71.44 & 0.1933\\
         \bottomrule

    \end{tabular}
    \label{tab:z1}
\end{table}

\subsection{Model}
In the previous subsection, we discussed extracting visual attributes for distortions. In this section, we use these attributes to fine-tune CLIP to identify distortions (Sub-Problem 1). Then, we train a regressor to estimate the quality score (Sub-Problem 2). Figure \ref{fig:2} shows our training approach. Overall, our training consists of two phases.

\subsubsection{Training Distortions}
Recalling from subsection \ref{subsec:PF}, consider dataset $\zeta = \left(\mathcal{I}, \mathcal{D}, \mathcal{P}, \mathcal{S}\right)$. As shown in figure \ref{fig:2} (Blocks 1 \& 2), we have following steps:
\begin{enumerate}
    \item (Image Encoding): Encode the image using the CLIP image encoder.
    \item (Predicting Distortion): First, for each attribute $a$ of each distortion $d\in \mathcal{D}$, we create positive and negative sentences.
    \begin{itemize}
        \item {\fontfamily{qcr}\selectfont There is <$a$> in the photo.}
        \item {\fontfamily{qcr}\selectfont There is not <$a$> in the photo.}
    \end{itemize}
    Then, compute the probability of attribute $a$ via computing dot product with image embedding and applying softmax function, i.e.,
    \begin{equation}
           \hat{P}(a|I) = \frac{e^{\mathbf{E}_t(a)^T \mathbf{E}_I(I)}}{e^{\mathbf{E}_{t}(a)^T \mathbf{E}_I(I)} + e^{\mathbf{E}_t(\tilde{a})^T \mathbf{E}_I(I)}} 
    \end{equation}
    where $\tilde{a}$ represent the negative prompt.
    Then, the distortion probability is computed as the weighted average of its corresponding attributes probability, i.e.,
    \begin{equation}
    \hat{P}(d|I) = \sum_{a \in \left\{\text{attributes of distortion }d\right\}} w_{a,d}\hat{P}(a|I)
    \end{equation}
    where ${w_{a,d}}$ is a learnable weight for contribution of attribute $a$ in identifying distortion $d$. Also we have $w_{a,d}\geq 0,\quad \sum_{a\in\{\text{attributes of distortion } d\}} w_{a,d}=1$.
\end{enumerate}
Finally, we use cross-entropy loss \cite{zhang2018generalized} for our distortion loss:
    \begin{align}
        &\mathcal{L}_{\text{dist}} = -\frac{1}{\left|\mathcal{D}\right|\left|\mathcal{I}\right|} \sum_{I\in \mathcal{I}} \sum_{d \in \mathcal{D}} \mathcal{P}(I,d)\log\left(\hat{P}(d|I)\right) \nonumber\\
         &-\frac{1}{\left|\mathcal{D}\right|\left|\mathcal{I}\right|} \sum_{I\in \mathcal{I}} \sum_{d \in \mathcal{D}}\left(1-\mathcal{P}(I,d)\right)\log \left(1-\hat{P}(d|I)\right)
    \end{align}

\subsubsection{Training Regressor}

Despite some prior works that use a discrete score model, here, we don't limit ourselves and use a regressor consisting of two hidden layers of MLP \cite{lecun2015deep} with SELU \cite{klambauer2017self} as the activation function.

As shown in figure \ref{fig:2} (Block 3), After fine-tuning CLIP with loss $\mathcal{L}_{\text{dist}}$, we freeze CLIP and retrieve all attributes\footnote{We don't use distortion probabilities because due to averaging, they have less information than attributes.} probabilities as the input features to the regressor. Then, we train the regressor with the following MSE loss \cite{sara2019image}:
\begin{equation}
    \mathcal{L}_{\text{reg}} = \frac{1}{\left|\mathcal{I}\right|} \sum_{I\in \mathcal{I}} \left(\text{MLP}(\{\hat{P}(a|I)\}) - \mathcal{S}(I)\right)^2
\end{equation}
where $\{\hat{P}(a|I)\}$ are all attributes probabilities retrieved from freezed CLIP.

In our approach, we used attribute probabilities as input features rather than image embeddings or other textual embeddings. This ensures that our regressor is trained solely on relevant features, making our model more transparent than previous methods. Additionally, this approach requires far fewer parameters for training on other datasets (due to the reduction in the number of input features), making it more computationally efficient.

\section{Experiments}
\subsection{Multi-Distortions Dataset}
Our model, as we discussed in section \ref{sec:method}, considers multiple distortions for each image. But current synthetic datasets such as KADID-10K \cite{kadid10k, deepfl-iqa}, TID2013 \cite{tid2013} and CSIQ \cite{csiq} only consider one distortion per image, which limits the further improvements. To this end, we generate a dataset of $100$k images, and each image can have any number of distortions. Below, we describe our procedure:
\begin{enumerate}
    \item We randomly select $10,000$ pristine images from KADIS-700k \cite{kadid10k}.
    \item We consider distortions of \cite{kadid10k} as our distortions set, i.e, $\mathcal{D}$. (although any other distortion set could be used)
    \item Repeat $10$ times:
    \begin{enumerate}
        \item For each image $I$, an integer $K_I$ in the range $\left[1,\left|\mathcal{D}\right|\right]$ is randomly chosen. These integers indicate the number of distortions to apply.
        \item For each image $I$ we select $K_I$ distortions from set $\mathcal{D}$ without replacement.
        \item After choosing types of distortions, we randomly choose their strengths, which are $5$ levels in our case.
        \item Distortions with their strengths are applied on corresponding images. 
        \end{enumerate}
\end{enumerate}
 The code for generating this dataset is written in Python and will be available upon publication. 

\begin{table*}[t]
    \small
    \centering
     \caption{Zero-Shot performance across different settings. In each row, the best result is bolded, and the second best is underlined. "-" means the dataset can not be evaluated. We achieved SOTA performance across all setups except for KADID-10k $\rightarrow$ LIVE.}
    \begin{tabular}{*{7}{c}}
        \toprule
            & & \multicolumn{5}{c}{Method}\\
            \cmidrule(lr){3-7}
         \begin{tabular}{l} Training \\ Dataset \end{tabular} & 
         \begin{tabular}{l} Testing \\ Dataset \end{tabular} & HyperIQA & CONTRIQUE & Re-IQA & ARNIQA & ExIQA (ours) \\
         \midrule
         KADID-10k &  TID2013 & 0.706 & 0.612 & \underline{0.777} & 0.760
         & \textbf{0.817}\\
         KADID-10k & CSIQ &  0.809 & 0.773 & 0.855 & \underline{0.882} & \textbf{0.913}\\
         KADID-10k & LVQ &  0.854 & 0.867 & 0.927 & \underline{0.933} & \textbf{0.966}\\
         KADID-10k & LIVE & \textbf{0.908} & \underline{0.900} & 0.892 & 0.898 & 0.883\\
         \midrule
         TID2013 & KADID-10k & 0.581 & 0.640 & 0.636 & \underline{0.726} & \textbf{0.845}\\
         TID2013 & CSIQ &  0.709 & 0.811 & 0.850 & 0.\underline{866} & \textbf{0.9154}\\
         TID2013 & LVQ & 0.844 & 0.893 & \underline{0.922} & 0.920 & \textbf{0.9562}\\
         TID2013 & LIVE & 0.876 & \underline{0.904} & 0.900 & 0.869 & \textbf{0.914}\\
         \midrule
         ExIQA & KADID-10k & - & - & - & - & \textbf{0.931}\\
         ExIQA & TID2013 & - & - & - & - & \textbf{0.8237}\\
         ExIQA & CSIQ & - & - & - & - & \textbf{0.939}\\
         ExIQA & LVQ & - & - & - & - & \textbf{0.940}\\
         ExIQA & LIVE & - & - & - & - & \textbf{0.922}\\
         \bottomrule
    \end{tabular}
    \label{tab:z2}
\end{table*}    

\subsection{Experimental Setup}
\subsubsection{Datasets}
Alongside our multi-distortion dataset, we also consider single distortions datasets such as KADID-10k \cite{kadid10k,lin2020deepfl}, TID2013 \cite{tid2013}, CSIQ \cite{csiq}, LIVE \cite{live} and LVQ \cite{lvq1, lvq2, lvq3} datasets.
LVQ dataset has videos instead of images, so we extract video frames and use them as the input images for our model. Tabel \ref{tabel:ds} shows dataset statistics.

\subsubsection{Hyperparametrs}
The number of attributes per distortion is set to five, i.e., $k=5$. Asking for more attributes increases redundancy.

To maintain generalization and computational efficiency, we use shallow-prompt tuning for datasets KADID-10k, TID2013, CSIQ, LIVE, and LVQ, using different learnable prompts due to the size of these datasets. Table \ref{tabel:m1} shows the prompt length we consider for each dataset. For the ExIQA (ours) dataset, we found out that fine-tuning the whole CLIP for a few epochs is more efficient than utilizing deep-prompt tuning due to the large size of this dataset.

For fine-tuning CLIP, we use the CosineLR scheduler with linear warmup \cite{cosineLR}. The maximum learning rate is set to 0.002, with a total of 100 epochs, of which 20 epochs are reserved for warmup. For the ExIQA dataset, we fine-tune over $5$ epochs with a learning rate of 5e-5. The regressor is trained for 100 epochs using the CosineLR scheduler. Additionally, we apply dropout \cite{srivastava2014dropout} with a probability of 0.2.

\subsubsection{Metrics \& Evaluation}
Our Evaluation consists of two parts. In the first phase, the capability of CLIP in recognizing distortions is tested through the accuracy of identifying ground-truth distortions and RMSE of the strength of the distortions.

Because the ground-truth strength of distortions is in discrete levels, we divide the interval $[0,1]$ into non-overlapping intervals. For example, if a dataset has $5$ ($+1$ for clean image) level of distortions, then our intervals become $[0, 0.1), [0.1, 0.3), [0.3, 0.5), [0.5, 0.7), [0.7, 0.9), [0.9, 1)$ where $0, 0.2, 0.4, 0.6, 0.8, 1$ are the ground-truth levels. Then, we check if our prediction for each distortion strength lies in the corresponding interval with the center of the ground-truth level. In this manner, we measure the accuracy of the model prediction as follows:
\begin{align}
    &\text{Accuracy} =\nonumber\\
    &\frac{1}{\left|\mathcal{I}\right|\left|\mathcal{D}\right|} \sum_{I\in\mathcal{I}} \sum_{d\in \mathcal{D}} \mathbb{I}\left[\hat{P}(d|I) \in \text{interval with center }\mathcal{P}(I,d)\right]
\end{align}
where $\mathbb{I}(.)$ is the indicator function and its value is $1$, if the condition at its argument is true, otherwise $0$.

For the first and last intervals, the ground-truth strength is on the edge of intervals, i.e., for the first interval is $0$, and for the last interval is $1$.

Also, we measure the RMSE error with
\begin{align}
    \text{RMSE} = \sqrt{\frac{1}{\left|\mathcal{I}\right|\left|\mathcal{D}\right|} \sum_{I\in \mathcal{I}}\sum_{d\in \mathcal{D}} \left(\hat{P}(d|I)-\mathcal{P}(d,I)\right)^2}
\end{align}

In the second phase, we evaluate our regressor in estimating the ground-truth quality scores of images through two standard metrics, \emph{PLCC} (\emph{P}earson \emph{L}inear \emph{C}orrelation \emph{C}oefficient) \cite{sedgwick2012pearson} and \emph{SRCC} (\emph{S}pearman's \emph{R}ank \emph{C}orrelation \emph{C}oefficient) \cite{sedgwick2014spearman}. In these correlation metrics, as close to one as we are, we have a better estimation.

\begin{figure*}[t]
    \centering
    \includegraphics[width=0.8\linewidth]{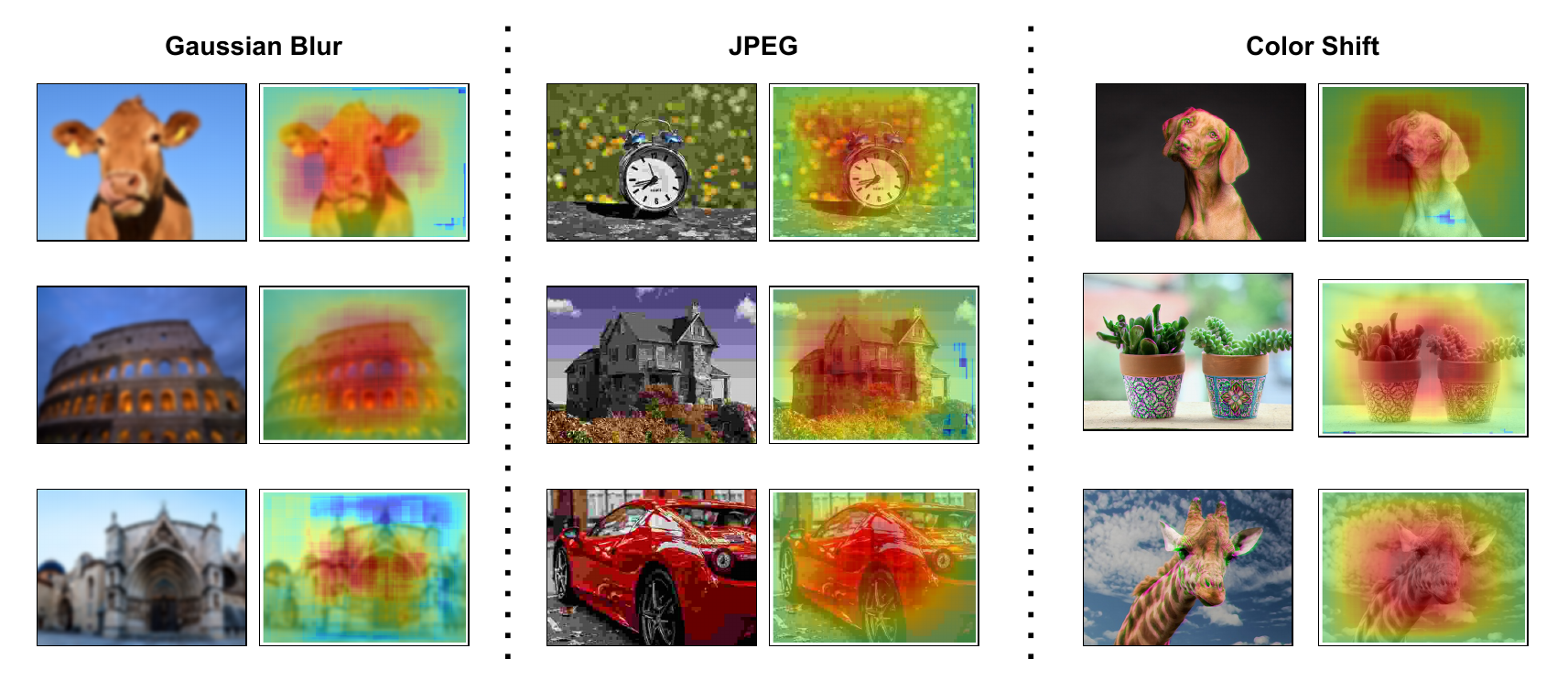}
    \caption{Saliency Maps.}
    \label{fig:3}
\end{figure*}

\subsection{Main Results}
We evaluate our method both on distortion identification and quality score estimation. Because of the lack of results for the distortion identification task, we can't compare our method with any baseline on this task. 

Table \ref{tabel:m1} shows the results on distortion identifications. For datasets with single distortion, we achieved high accuracy (average $95.77\%$) and low RMSE (average $0.0502$) with only training a few prompts. This shows the CLIP capability for distortion detection, which was not previously studied.
The result in the multi-distortion dataset, ExIQA (ours), is worse than that in single-distortion datasets. When multiple distortions are sequentially applied to a photo, the distinction between them is much harder than when only one is used because the distortions can affect each other. To this end, we still have accuracy near 70\% and RMSE of $0.156$, which is reasonable considering the task's difficulty.

Table \ref{tabel:m2} shows the results on datasets KADID-10k, TID2013, CSIQ, LVQ, and LIVE. For each dataset, we fine-tune CLIP with distortions on that dataset and train regressor with attribute probabilities retrieved from fine-tuned CLIP. As can be seen, we achieved the best results in these five datasets (in the PLCC columns of TID2013 and CSIQ, we got the second-best results). This shows that our approach, besides its transparency in detecting distortions and choosing regressor features, can compete with other baselines and achieve SOTA performance in multiple datasets.

\subsection{Zero-Shot Reuslts}
To show the generalizability of our model, we conducted zero-shot experiments. 

Tabel \ref{tab:z1} shows the zero-shot results on the distortion identification task. The leading cause for the drop in performance is different distortion level scales across different datasets. This is more obvious when the testing datasets are CSIQ and LIVE (accuracy is around 70\%), where scales of distortions are different from KADID-10k, TID2013, LVQ, and ExIQA. However, the performance is still promising for other datasets, especially when the training dataset is ExIQA (ours) with the effect of multi-distortion training on model representation. Also, the results are good when the test dataset is too specific and belongs to a different domain, such as LVQ, which contains Laparoscopic videos \cite{lvq1}.

Tabel \ref{tab:z2} shows the zero-shot results of the quality score prediction based on the SRCC metric. Our setup is as follows: We fine-tune CLIP on a training dataset (first column), then freeze it and retrieve attribute probabilities of the training dataset for fine-tuning a regressor on a testing dataset (column 2). For other baselines, we follow their setup on corresponding papers. As the results show, our method outperforms other baselines in almost all setups (except KADID-10k $\rightarrow$ LIVE). Because our dataset, ExIQA, is specialized for distortion identification tasks. So, we don't evaluate zero-shot results on other baselines. However, compared to other training setups, ExIQA produces better features (because of multi-distortion training) and gives higher SRCC on test datasets.

\subsection{Saliency Maps}
To visualize the distortion identification process, we provide saliency maps in figure \ref{fig:3}. Because distortions affect all areas of an image, to better understand the model decision-making, we consider photos with approximately constant background to check that the model looks at the correct parts of an image (like edges). Also, for readability and simplification of visualization, we consider single distorted photos. As shown in figure \ref{fig:3}, our training approach causes CLIP to focus on the most relevant and distorted parts of the image, increasing its explainability and accuracy.

\section{Conclusion}
In this paper, we examine BIQA from a distortion identification perspective and propose an approach via VLMs such as CLIP for identifying distortion and predicting quality scores in an explainable manner. For the first objective, explainable distortion identification, we extracted attributes or effects of distortions and utilized them instead of using the names of distortions, which can be ambiguous and limits the scalability to more distortions. We also generated a 100,000 multi-distorted dataset to improve our training. In the regressor network, we used attribute probabilities as input features, which boosts the explainability of the regressor, too. Overall, we achieved SOTA performance across different datasets and domains. As a future work, we want to scale our method to authentically distorted images, which probably contain more complicated distortions that can not be found in synthetic datasets. Because our approach relies on attributes rather than distortion names, it has the potential to be developed for authentically distorted images, too.

\clearpage
\bibliographystyle{IEEEtran}
\bibliography{ExIQA}

\begin{thebibliography}{10}
\providecommand{\url}[1]{#1}
\csname url@samestyle\endcsname
\providecommand{\newblock}{\relax}
\providecommand{\bibinfo}[2]{#2}
\providecommand{\BIBentrySTDinterwordspacing}{\spaceskip=0pt\relax}
\providecommand{\BIBentryALTinterwordstretchfactor}{4}
\providecommand{\BIBentryALTinterwordspacing}{\spaceskip=\fontdimen2\font plus
\BIBentryALTinterwordstretchfactor\fontdimen3\font minus \fontdimen4\font\relax}
\providecommand{\BIBforeignlanguage}[2]{{%
\expandafter\ifx\csname l@#1\endcsname\relax
\typeout{** WARNING: IEEEtran.bst: No hyphenation pattern has been}%
\typeout{** loaded for the language `#1'. Using the pattern for}%
\typeout{** the default language instead.}%
\else
\language=\csname l@#1\endcsname
\fi
#2}}
\providecommand{\BIBdecl}{\relax}
\BIBdecl

\bibitem{wang2006modern}
Z.~Wang and A.~C. Bovik, ``Modern image quality assessment,'' Ph.D. dissertation, Springer, 2006.

\bibitem{zhai2020perceptual}
G.~Zhai and X.~Min, ``Perceptual image quality assessment: a survey,'' \emph{Science China Information Sciences}, vol.~63, pp. 1--52, 2020.

\bibitem{bosse2017deep}
S.~Bosse, D.~Maniry, K.-R. M{\"u}ller, T.~Wiegand, and W.~Samek, ``Deep neural networks for no-reference and full-reference image quality assessment,'' \emph{IEEE Transactions on image processing}, vol.~27, no.~1, pp. 206--219, 2017.

\bibitem{athar2019comprehensive}
S.~Athar and Z.~Wang, ``A comprehensive performance evaluation of image quality assessment algorithms,'' \emph{Ieee Access}, vol.~7, pp. 140\,030--140\,070, 2019.

\bibitem{chow2016review}
L.~S. Chow and R.~Paramesran, ``Review of medical image quality assessment,'' \emph{Biomedical signal processing and control}, vol.~27, pp. 145--154, 2016.

\bibitem{liang2020deep}
D.~Liang, X.~Gao, W.~Lu, and L.~He, ``Deep multi-label learning for image distortion identification,'' \emph{Signal processing}, vol. 172, p. 107536, 2020.

\bibitem{yan2021precise}
C.~Yan, T.~Teng, Y.~Liu, Y.~Zhang, H.~Wang, and X.~Ji, ``Precise no-reference image quality evaluation based on distortion identification,'' \emph{ACM Transactions on Multimedia Computing, Communications, and Applications (TOMM)}, vol.~17, no.~3s, pp. 1--21, 2021.

\bibitem{zhang2024vision}
J.~Zhang, J.~Huang, S.~Jin, and S.~Lu, ``Vision-language models for vision tasks: A survey,'' \emph{IEEE Transactions on Pattern Analysis and Machine Intelligence}, 2024.

\bibitem{zhang2021vinvl}
P.~Zhang, X.~Li, X.~Hu, J.~Yang, L.~Zhang, L.~Wang, Y.~Choi, and J.~Gao, ``Vinvl: Revisiting visual representations in vision-language models,'' in \emph{Proceedings of the IEEE/CVF conference on computer vision and pattern recognition}, 2021, pp. 5579--5588.

\bibitem{radford2021learning}
A.~Radford, J.~W. Kim, C.~Hallacy, A.~Ramesh, G.~Goh, S.~Agarwal, G.~Sastry, A.~Askell, P.~Mishkin, J.~Clark \emph{et~al.}, ``Learning transferable visual models from natural language supervision,'' in \emph{International conference on machine learning}.\hskip 1em plus 0.5em minus 0.4em\relax PMLR, 2021, pp. 8748--8763.

\bibitem{arniqa}
L.~Agnolucci, L.~Galteri, M.~Bertini, and A.~Del~Bimbo, ``Arniqa: Learning distortion manifold for image quality assessment,'' in \emph{Proceedings of the IEEE/CVF Winter Conference on Applications of Computer Vision}, 2024, pp. 189--198.

\bibitem{chen2020simple}
T.~Chen, S.~Kornblith, M.~Norouzi, and G.~Hinton, ``A simple framework for contrastive learning of visual representations,'' in \emph{International conference on machine learning}.\hskip 1em plus 0.5em minus 0.4em\relax PMLR, 2020, pp. 1597--1607.

\bibitem{ma2023rectify}
C.~Ma, L.~Zhao, Y.~Chen, L.~Guo, T.~Zhang, X.~Hu, D.~Shen, X.~Jiang, and T.~Liu, ``Rectify vit shortcut learning by visual saliency,'' \emph{IEEE Transactions on Neural Networks and Learning Systems}, 2023.

\bibitem{zhang2023blind}
W.~Zhang, G.~Zhai, Y.~Wei, X.~Yang, and K.~Ma, ``Blind image quality assessment via vision-language correspondence: A multitask learning perspective,'' in \emph{Proceedings of the IEEE/CVF conference on computer vision and pattern recognition}, 2023, pp. 14\,071--14\,081.

\bibitem{zhao2023survey}
W.~X. Zhao, K.~Zhou, J.~Li, T.~Tang, X.~Wang, Y.~Hou, Y.~Min, B.~Zhang, J.~Zhang, Z.~Dong \emph{et~al.}, ``A survey of large language models,'' \emph{arXiv preprint arXiv:2303.18223}, 2023.

\bibitem{brown2020language}
T.~B. Brown, ``Language models are few-shot learners,'' \emph{arXiv preprint arXiv:2005.14165}, 2020.

\bibitem{kadid10k}
H.~Lin, V.~Hosu, and D.~Saupe, ``Kadid-10k: A large-scale artificially distorted iqa database,'' in \emph{2019 Tenth International Conference on Quality of Multimedia Experience (QoMEX)}.\hskip 1em plus 0.5em minus 0.4em\relax IEEE, 2019, pp. 1--3.

\bibitem{niqe}
A.~Mittal, R.~Soundararajan, and A.~C. Bovik, ``Making a “completely blind” image quality analyzer,'' \emph{IEEE Signal processing letters}, vol.~20, no.~3, pp. 209--212, 2012.

\bibitem{brisque}
A.~Mittal, A.~K. Moorthy, and A.~C. Bovik, ``No-reference image quality assessment in the spatial domain,'' \emph{IEEE Transactions on image processing}, vol.~21, no.~12, pp. 4695--4708, 2012.

\bibitem{fu2018quality}
Z.~Fu, F.~Shao, Q.~Jiang, R.~Fu, and Y.-S. Ho, ``Quality assessment of retargeted images using hand-crafted and deep-learned features,'' \emph{IEEE Access}, vol.~6, pp. 12\,008--12\,018, 2018.

\bibitem{reinagel1999natural}
P.~Reinagel and A.~M. Zador, ``Natural scene statistics at the centre of gaze,'' \emph{Network: Computation in Neural Systems}, vol.~10, no.~4, p. 341, 1999.

\bibitem{ye2012no}
P.~Ye and D.~Doermann, ``No-reference image quality assessment using visual codebooks,'' \emph{IEEE Transactions on Image Processing}, vol.~21, no.~7, pp. 3129--3138, 2012.

\bibitem{cornia}
P.~Ye, J.~Kumar, L.~Kang, and D.~Doermann, ``Unsupervised feature learning framework for no-reference image quality assessment,'' in \emph{2012 IEEE conference on computer vision and pattern recognition}.\hskip 1em plus 0.5em minus 0.4em\relax IEEE, 2012, pp. 1098--1105.

\bibitem{hosa}
J.~Xu, P.~Ye, Q.~Li, H.~Du, Y.~Liu, and D.~Doermann, ``Blind image quality assessment based on high order statistics aggregation,'' \emph{IEEE Transactions on Image Processing}, vol.~25, no.~9, pp. 4444--4457, 2016.

\bibitem{dbcnn}
W.~Zhang, K.~Ma, J.~Yan, D.~Deng, and Z.~Wang, ``Blind image quality assessment using a deep bilinear convolutional neural network,'' \emph{IEEE Transactions on Circuits and Systems for Video Technology}, vol.~30, no.~1, pp. 36--47, 2020.

\bibitem{lin2020deepfl}
H.~Lin, V.~Hosu, and D.~Saupe, ``Deepfl-iqa: Weak supervision for deep iqa feature learning,'' \emph{arXiv preprint arXiv:2001.08113}, 2020.

\bibitem{bianco2018use}
S.~Bianco, L.~Celona, P.~Napoletano, and R.~Schettini, ``On the use of deep learning for blind image quality assessment,'' \emph{Signal, Image and Video Processing}, vol.~12, pp. 355--362, 2018.

\bibitem{angelov2020towards}
P.~Angelov and E.~Soares, ``Towards explainable deep neural networks (xdnn),'' \emph{Neural Networks}, vol. 130, pp. 185--194, 2020.

\bibitem{zhang2019deep}
Y.~Zhang, B.~Peng, X.~Zhou, C.~Xiang, and D.~Wang, ``A deep convolutional neural network for topology optimization with strong generalization ability,'' \emph{arXiv preprint arXiv:1901.07761}, 2019.

\bibitem{reiqa}
A.~Saha, S.~Mishra, and A.~C. Bovik, ``Re-iqa: Unsupervised learning for image quality assessment in the wild,'' in \emph{Proceedings of the IEEE/CVF conference on computer vision and pattern recognition}, 2023, pp. 5846--5855.

\bibitem{contrique}
P.~C. Madhusudana, N.~Birkbeck, Y.~Wang, B.~Adsumilli, and A.~C. Bovik, ``Image quality assessment using contrastive learning,'' \emph{IEEE Transactions on Image Processing}, vol.~31, pp. 4149--4161, 2022.

\bibitem{zhou2023collaborative}
Z.~Zhou, F.~Zhou, and G.~Qiu, ``Collaborative auto-encoding for blind image quality assessment,'' in \emph{2023 IEEE International Conference on Multimedia and Expo (ICME)}.\hskip 1em plus 0.5em minus 0.4em\relax IEEE, 2023, pp. 1295--1300.

\bibitem{mao2023doubly}
C.~Mao, R.~Teotia, A.~Sundar, S.~Menon, J.~Yang, X.~Wang, and C.~Vondrick, ``Doubly right object recognition: A why prompt for visual rationales,'' in \emph{Proceedings of the IEEE/CVF Conference on Computer Vision and Pattern Recognition}, 2023, pp. 2722--2732.

\bibitem{rasekh2024ecor}
A.~Rasekh, S.~K. Ranjbar, M.~Heidari, and W.~Nejdl, ``Ecor: Explainable clip for object recognition,'' \emph{arXiv preprint arXiv:2404.12839}, 2024.

\bibitem{zellers2019recognition}
R.~Zellers, Y.~Bisk, A.~Farhadi, and Y.~Choi, ``From recognition to cognition: Visual commonsense reasoning,'' in \emph{Proceedings of the IEEE/CVF conference on computer vision and pattern recognition}, 2019, pp. 6720--6731.

\bibitem{davis2015commonsense}
E.~Davis and G.~Marcus, ``Commonsense reasoning and commonsense knowledge in artificial intelligence,'' \emph{Communications of the ACM}, vol.~58, no.~9, pp. 92--103, 2015.

\bibitem{jia2022visual}
M.~Jia, L.~Tang, B.-C. Chen, C.~Cardie, S.~Belongie, B.~Hariharan, and S.-N. Lim, ``Visual prompt tuning,'' in \emph{European Conference on Computer Vision}.\hskip 1em plus 0.5em minus 0.4em\relax Springer, 2022, pp. 709--727.

\bibitem{vaswani2017attention}
A.~Vaswani, ``Attention is all you need,'' \emph{Advances in Neural Information Processing Systems}, 2017.

\bibitem{ba2016layer}
J.~Ba, ``Layer normalization,'' \emph{arXiv preprint arXiv:1607.06450}, 2016.

\bibitem{lecun2015deep}
Y.~LeCun, Y.~Bengio, and G.~Hinton, ``Deep learning,'' \emph{nature}, vol. 521, no. 7553, pp. 436--444, 2015.

\bibitem{tid2013}
N.~Ponomarenko, L.~Jin, O.~Ieremeiev, V.~Lukin, K.~Egiazarian, J.~Astola, B.~Vozel, K.~Chehdi, M.~Carli, F.~Battisti \emph{et~al.}, ``Image database tid2013: Peculiarities, results and perspectives,'' \emph{Signal processing: Image communication}, vol.~30, pp. 57--77, 2015.

\bibitem{csiq}
E.~C. Larson and D.~M. Chandler, ``Most apparent distortion: full-reference image quality assessment and the role of strategy,'' \emph{Journal of electronic imaging}, vol.~19, no.~1, pp. 011\,006--011\,006, 2010.

\bibitem{live}
H.~R. Sheikh, M.~F. Sabir, and A.~C. Bovik, ``A statistical evaluation of recent full reference image quality assessment algorithms,'' \emph{IEEE Transactions on image processing}, vol.~15, no.~11, pp. 3440--3451, 2006.

\bibitem{lvq1}
Z.~A. Khan, A.~Beghdadi, M.~Kaaniche, F.~Alaya-Cheikh, and O.~Gharbi, ``A neural network based framework for effective laparoscopic video quality assessment,'' \emph{Computerized Medical Imaging and Graphics}, vol. 101, p. 102121, 2022.

\bibitem{espejel2023gpt}
J.~L. Espejel, E.~H. Ettifouri, M.~S.~Y. Alassan, E.~M. Chouham, and W.~Dahhane, ``Gpt-3.5, gpt-4, or bard? evaluating llms reasoning ability in zero-shot setting and performance boosting through prompts,'' \emph{Natural Language Processing Journal}, vol.~5, p. 100032, 2023.

\bibitem{hyperiqa}
S.~Su, Q.~Yan, Y.~Zhu, C.~Zhang, X.~Ge, J.~Sun, and Y.~Zhang, ``Blindly assess image quality in the wild guided by a self-adaptive hyper network,'' in \emph{Proceedings of the IEEE/CVF conference on computer vision and pattern recognition}, 2020, pp. 3667--3676.

\bibitem{tres}
S.~A. Golestaneh, S.~Dadsetan, and K.~M. Kitani, ``No-reference image quality assessment via transformers, relative ranking, and self-consistency,'' in \emph{Proceedings of the IEEE/CVF winter conference on applications of computer vision}, 2022, pp. 1220--1230.

\bibitem{zhang2018generalized}
Z.~Zhang and M.~Sabuncu, ``Generalized cross entropy loss for training deep neural networks with noisy labels,'' \emph{Advances in neural information processing systems}, vol.~31, 2018.

\bibitem{klambauer2017self}
G.~Klambauer, T.~Unterthiner, A.~Mayr, and S.~Hochreiter, ``Self-normalizing neural networks,'' \emph{Advances in neural information processing systems}, vol.~30, 2017.

\bibitem{sara2019image}
U.~Sara, M.~Akter, and M.~S. Uddin, ``Image quality assessment through fsim, ssim, mse and psnr—a comparative study,'' \emph{Journal of Computer and Communications}, vol.~7, no.~3, pp. 8--18, 2019.

\bibitem{deepfl-iqa}
H.~Lin, V.~Hosu, and D.~Saupe, ``Deepfl-iqa: Weak supervision for deep iqa feature learning,'' \emph{arXiv preprint arXiv:2001.08113}, 2020.

\bibitem{lvq2}
Z.~A. Khan, A.~Beghdadi, F.~A. Cheikh, M.~Kaaniche, E.~Pelanis, R.~Palomar, {\AA}.~A. Fretland, B.~Edwin, and O.~J. Elle, ``Towards a video quality assessment based framework for enhancement of laparoscopic videos,'' in \emph{Medical Imaging 2020: Image Perception, Observer Performance, and Technology Assessment}, vol. 11316.\hskip 1em plus 0.5em minus 0.4em\relax SPIE, 2020, pp. 129--136.

\bibitem{lvq3}
Z.~A. Khan, A.~Beghdadi, M.~Kaaniche, and F.~A. Cheikh, ``Residual networks based distortion classification and ranking for laparoscopic image quality assessment,'' in \emph{2020 IEEE International Conference on Image Processing (ICIP)}.\hskip 1em plus 0.5em minus 0.4em\relax IEEE, 2020, pp. 176--180.

\bibitem{cosineLR}
I.~Loshchilov and F.~Hutter, ``Sgdr: Stochastic gradient descent with warm restarts,'' \emph{arXiv preprint arXiv:1608.03983}, 2016.

\bibitem{srivastava2014dropout}
N.~Srivastava, G.~Hinton, A.~Krizhevsky, I.~Sutskever, and R.~Salakhutdinov, ``Dropout: a simple way to prevent neural networks from overfitting,'' \emph{The journal of machine learning research}, vol.~15, no.~1, pp. 1929--1958, 2014.

\bibitem{sedgwick2012pearson}
P.~Sedgwick, ``Pearson’s correlation coefficient,'' \emph{Bmj}, vol. 345, 2012.

\bibitem{sedgwick2014spearman}
------, ``Spearman’s rank correlation coefficient,'' \emph{Bmj}, vol. 349, 2014.

\end{thebibliography}

\end{document}